\documentclass[letterpaper]{article} 
\usepackage[preprint]{aaai2027}  
\usepackage[hyphens]{url}  
\usepackage{graphicx} 
\usepackage{natbib}  
\usepackage{caption} 
\usepackage{amsmath}
\usepackage{amssymb}
\usepackage{newfloat}
\usepackage{listings}
\usepackage{multirow}
\usepackage{makecell}
\usepackage{xcolor}
\definecolor{linkblue}{RGB}{80,140,200}
\DeclareCaptionStyle{ruled}{labelfont=normalfont,labelsep=colon,strut=off} 
\usepackage{booktabs}
\title{SewFusion: Tailored Generation of Topology and Panel-Level Geometry for Sewing Patterns}
\author{
    Jiaxin Lin\textsuperscript{\rm 1},
    Xiao Pan\textsuperscript{\rm 1},
    Hangjie Yuan\textsuperscript{\rm 2},
    Luyan Liang\textsuperscript{\rm 1},
    Wan Li\textsuperscript{\rm 1},
    Daquan Feng\textsuperscript{\rm 1}
}
\affiliations{
    \textsuperscript{\rm 1}Digital Creative Technology Laboratory, Shenzhen University\\
    \textsuperscript{\rm 2}Zhejiang University\\
    \{linjiaxin2022,liwan2022,liangluyan\}@email.szu.edu.cn,
    \{xiaopan97,fdquan\}@szu.edu.cn,
    hj.yuan@zju.edu.cn
}

\begin{document}
\maketitle

\begin{abstract}
Generating sewing patterns from images and text requires modeling a heterogeneous representation composed of discrete topology and continuous geometry. Existing methods mainly follow two paradigms: diffusion-based methods enable holistic geometry generation by converting the entire pattern into a continuous representation, but weaken discrete topology modeling; in contrast, autoregressive methods preserve discrete topology through next-token prediction, but tie continuous geometry regression to token-level hidden states with limited panel-level context. To bridge this gap, we propose SewFusion, a unified autoregressive framework that adopts tailored generation mechanisms for discrete topology and panel-level continuous geometry, using next-token prediction for the former and flow matching for the latter. To support panel-level continuous geometry generation, we introduce a Panel Geometry VAE that learns a fixed-size latent space for variable-length panel geometry, together with Panel Geometry Flow for latent generation. We further propose Panel-Forcing to reduce the training--inference mismatch in topology context and improve robustness to topology prediction errors. Extensive experiments on SewFactory and GCD-MM demonstrate that SewFusion consistently outperforms previous state-of-the-art methods across various settings, achieving $+6.36\%$ Panel Accuracy, $+11.30\%$ Stitch Accuracy, and $-1.90$ Vertex L2 error in the image-text-based generation setting. The project page is available at {\textcolor{linkblue}{\textit{https://jason96a.github.io/SewFusion/}}}.
\end{abstract}


\section{Introduction}

\begin{figure}[t]
    \centering
    \setlength{\abovecaptionskip}{0pt}
    \includegraphics[width=\columnwidth]{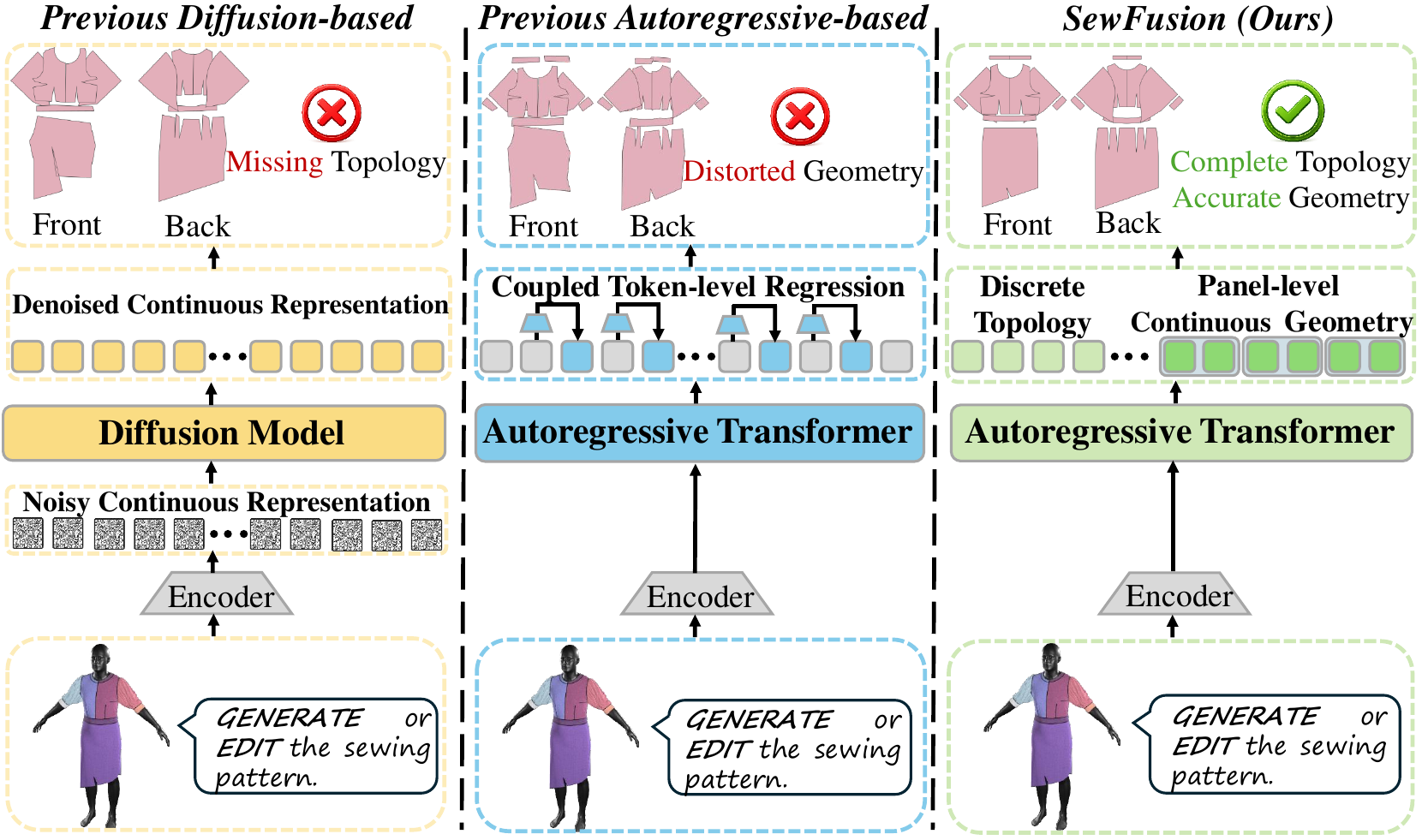}
    \caption{\textbf{Comparison between existing paradigms and SewFusion.} SewFusion enables tailored generation of discrete topology and panel-level continuous geometry within a unified autoregressive framework.}
    \label{fig:baseline_pipeline}
    \vspace{-4mm}
\end{figure}

Digital garment design plays a central role in virtual try-on, fashion design, digital humans, and physically based simulation~\cite{zeng2026dresswild,li2024garmentrecovery,wang2026seamcorrespondence}. At its core lies the sewing pattern, a heterogeneous garment representation that combines \textit{discrete topology}, including panel identities, edge types, and stitch relations, with \textit{continuous geometry}, including edge vertices, curve control parameters, and panel rigid transformations.
Designing sewing patterns manually in specialized CAD systems is labor-intensive and requires substantial expertise~\cite{li2025diffusionmapping,li2023isp,li2026patterngsl,riosnavarro2026autosew,li2024manipulated}. To automate this process, recent methods have explored learning-based sewing pattern prediction and generation from diverse visual, geometric, and textual inputs~\cite{yang2018physics,wang2018shared,korosteleva2022neuraltailor,liu2023sewformer,tatsukawa2025garmentimage,li2025dress,nakayama2026garment}.

As illustrated in Fig.~\ref{fig:baseline_pipeline}, recent generative approaches mainly follow two paradigms: \textit{Diffusion-based methods} encode the entire sewing pattern into a unified continuous representation~\cite{li2025garmentdiffusion,liu2025sewingldm,li2025garmagenet}. While effective at modeling continuous geometry, this formulation converts discrete topology, including panel existence, edge types, and stitch relations, into continuous prediction, often causing missing topology. \textit{In contrast, autoregressive-based methods}~\cite{nakayama2025aipparel} preserve discrete topology through next-token prediction, but regress continuous geometry from the corresponding token-level hidden states~\cite{he2024dresscode,nakayama2025aipparel}. Such token-level regression couples discrete topology and continuous geometry within shared hidden states and lacks the panel-level context needed to ensure boundary closure, ultimately leading to distorted geometry. \textit{Consequently, existing methods have yet to provide a framework that models topology discretely while enabling decoupled, panel-level continuous geometry generation.}

\begin{figure*}[t]
    \centering
    \setlength{\abovecaptionskip}{0pt}
    \includegraphics[width=0.8\textwidth]{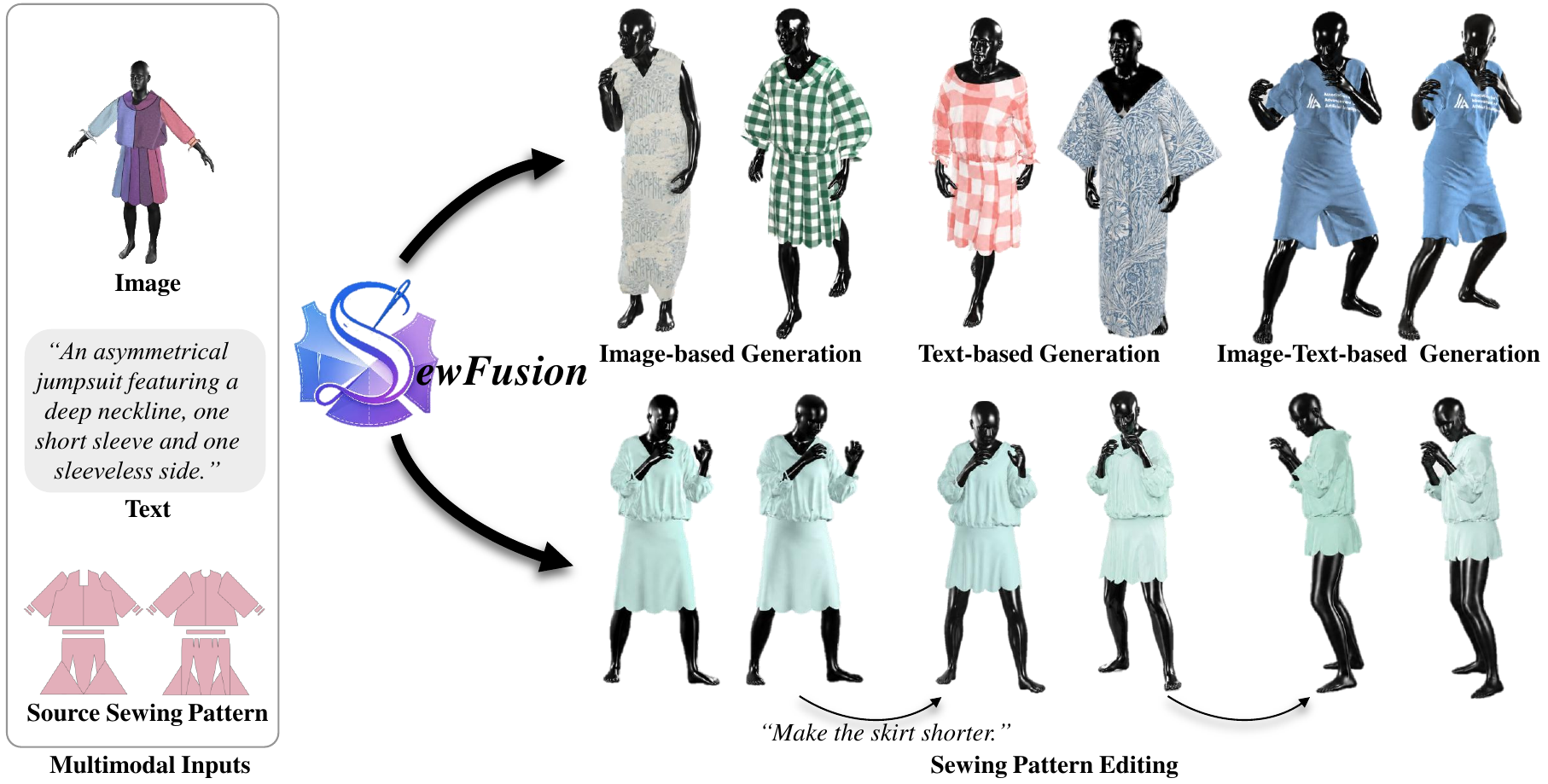}
    \caption{\textbf{SewFusion is a versatile framework} that generates simulation-ready sewing patterns from images, text, or their combination, and edits existing patterns according to user instructions. The resulting sewing patterns can be used in garment simulation software to assemble and simulate the corresponding 3D garments.}
    \label{fig:teaser}
    \vspace{-3mm}
\end{figure*}

To this end, we propose SewFusion, a unified autoregressive model that adopts tailored generation mechanisms for discrete topology and panel-level continuous geometry, using next-token prediction for the former and flow matching for the latter. To enable panel-level continuous geometry generation, we introduce a \textbf{Panel Geometry VAE} that encodes the variable-length geometry parameters of each panel into a fixed-size structured latent representation. These panel-level latent representations are generated by \textbf{Panel Geometry Flow} conditioned on the multimodal input and the preceding discrete topology, and are subsequently decoded into continuous geometry parameters. However, training Panel Geometry Flow with ground-truth topology creates a training–inference mismatch in topology context. To address this mismatch, we further introduce \textbf{Panel-Forcing}, which exposes Panel Geometry Flow to model-generated topology during training, thereby improving its robustness to topology prediction errors.

Our main contributions are summarized as follows:

\begin{itemize}
    \item We propose SewFusion, a unified autoregressive framework that adopts tailored generation mechanisms for discrete topology and panel-level continuous geometry, using next-token prediction for the former and flow matching for the latter.
    
    \item We introduce a Panel Geometry VAE and Panel Geometry Flow, enabling coherent generation of variable-length panel geometry in a fixed-size panel-level latent space. We further propose Panel-Forcing to reduce the training--inference mismatch in topology context and improve robustness to topology prediction errors.
    
    \item Extensive experiments on SewFactory and GCD-MM demonstrate consistent improvements across sewing pattern generation and editing settings. In image-text-based generation, SewFusion improves Panel Accuracy and Stitch Accuracy by \textbf{6.36\%} and \textbf{11.30\%}, respectively, while reducing Vertex L2 error by \textbf{1.90}.
\end{itemize}

\section{Related Work}

\subsection{Sewing Pattern Generation}
Early methods fit predefined templates or parameterized programs to observed garment appearance or geometry~\cite{yang2018physics,wang2018shared,bang2021estimating}. GarmentCode~\cite{korosteleva2023garmentcode} improves controllability and structural validity through hierarchical components and interpretable parameters, but fixed garment categories and topologies limit generalization to unseen panel and stitch structures~\cite{chi2021garmentnets,zhao2021anchorudf,zhu2022reef,deluigi2023drapenet,qiu2023recmv,li2024diffavatar}.

Recent diffusion-based methods model multimodal sewing pattern distributions. SewingLDM~\cite{liu2025sewingldm} performs conditional diffusion in a latent pattern space, GarmentDiffusion~\cite{li2025garmentdiffusion} jointly denoises tokenized pattern edges, and GarmageNet~\cite{li2025garmagenet} couples 2D patterns with 3D garment geometry. Although these approaches effectively model continuous geometry and diverse outputs, unifying panel existence, geometry, placement, and stitching in continuous representations can obscure discrete semantics and structural constraints.

Autoregressive methods instead serialize patterns into structured sequences. DressCode~\cite{he2024dresscode} quantizes pattern attributes into scalar tokens, while AIpparel~\cite{nakayama2025aipparel} supports image-, text-, and editing-conditioned generation through multimodal token prediction. Related methods generate structured descriptions or pattern-making programs~\cite{bian2025chatgarment,zhou2025design2garmentcode}. While autoregressive modeling naturally supports variable-length topology, representing continuous geometry through quantized tokens or token-level regression fragments coherent panel shapes and may cause topology--geometry ambiguity and error accumulation over long rollouts.

Existing generative methods either force all sewing-pattern attributes into a unified continuous space or discretize the entire pattern into a sequential token representation. Both paradigms overlook the inherent heterogeneity of sewing patterns, where topology is discrete but geometry is continuous. In contrast, our method assigns specialized generation mechanisms to the two modalities, using autoregressive prediction for discrete topology and topology-conditioned flow matching for continuous panel-level geometry.

\subsection{Self-forcing for Robust Generation}
Autoregressive models are commonly trained with teacher forcing, creating exposure bias because inference relies on model-generated rather than ground-truth prefixes. Recent methods reduce this gap by incorporating self-generated tokens or rollouts during training~\cite{cen2025bridging,huang2025self,liu2025rolling,guo2025resampling}. In visual generation, self-forcing and related approaches improve long-horizon robustness through inference-like rollouts, self-resampling, extended denoising, and causal distillation~\cite{yang2025longlive,zhang2025antiexposure,zhu2026causal}, while inference-time strategies reinforce prefix consistency~\cite{liao2026vpg}. These ideas are particularly important for structured generation, where early symbolic errors propagate to later predictions, motivating our approach to improve continuous geometry robustness.

\section{Preliminary}

\subsection{Problem Formulation}

Given a multimodal input $x$, consisting of a garment image, a textual description, or a source sewing pattern, our objective is to generate a simulation-ready sewing pattern $S$. We formulate sewing pattern generation as a heterogeneous discrete--continuous problem and decompose ${S}$ into discrete topology $T$ and panel-level continuous geometry $G$. The generation process is factorized as
\begin{equation}
    p({S}\mid x)
    =
    p_{\theta}(T\mid x)\,
    p_{\theta,\psi}(G\mid T,x),
    \label{eq:factorization}
\end{equation}
where the same causal multimodal Transformer with parameters $\theta$ generates discrete topology through next-token prediction and panel-level continuous geometry through flow matching. The Panel Geometry VAE with parameters $\psi$ defines the latent space in which continuous geometry is generated and decoded. This factorization models topology discretely while enabling decoupled continuous geometry generation at the panel level, thereby tailoring the generation mechanism to each representation.

\subsection{Sewing Pattern Representation}

\subsubsection{Discrete Topology Representation.}

A sewing pattern contains $N$ panels. For the $i$-th panel, its discrete topology $T_i$ specifies the panel identity, the ordered edge types, and the stitch tags associated with its edges. Edge types include line, quadratic B\'ezier curve, cubic B\'ezier curve, and circular arc. The complete topology is denoted by $T=\{T_i\}_{i=1}^{N}$, which determines how many panels the sewing pattern contains, how they are organized, what type each edge takes, and how edges are stitched. The associated continuous geometry is represented separately by $G$, as described next.

\subsubsection{Continuous Geometry Representation.}

For the $i$-th panel, $E_i$ represents the geometry of edges arranged sequentially along the panel boundary. Each edge geometry is composed of vertices and type-specific curve control parameters. $H_i$ represents the panel rigid transformation, consisting of a translation $t_i \in \mathbb{R}^{3}$ and a unit quaternion $q_i \in \mathbb{R}^{4}$ representing its rotation. Together, $E_i$ and $H_i$ constitute the continuous geometry $G_i$ of the $i$-th panel. The complete continuous geometry is denoted by $G=\{G_i\}_{i=1}^{N}$, which determines the shape and spatial placement of panels.

\begin{figure*}[t]
    \centering
    \setlength{\abovecaptionskip}{0pt}
    \includegraphics[width=0.90\textwidth]{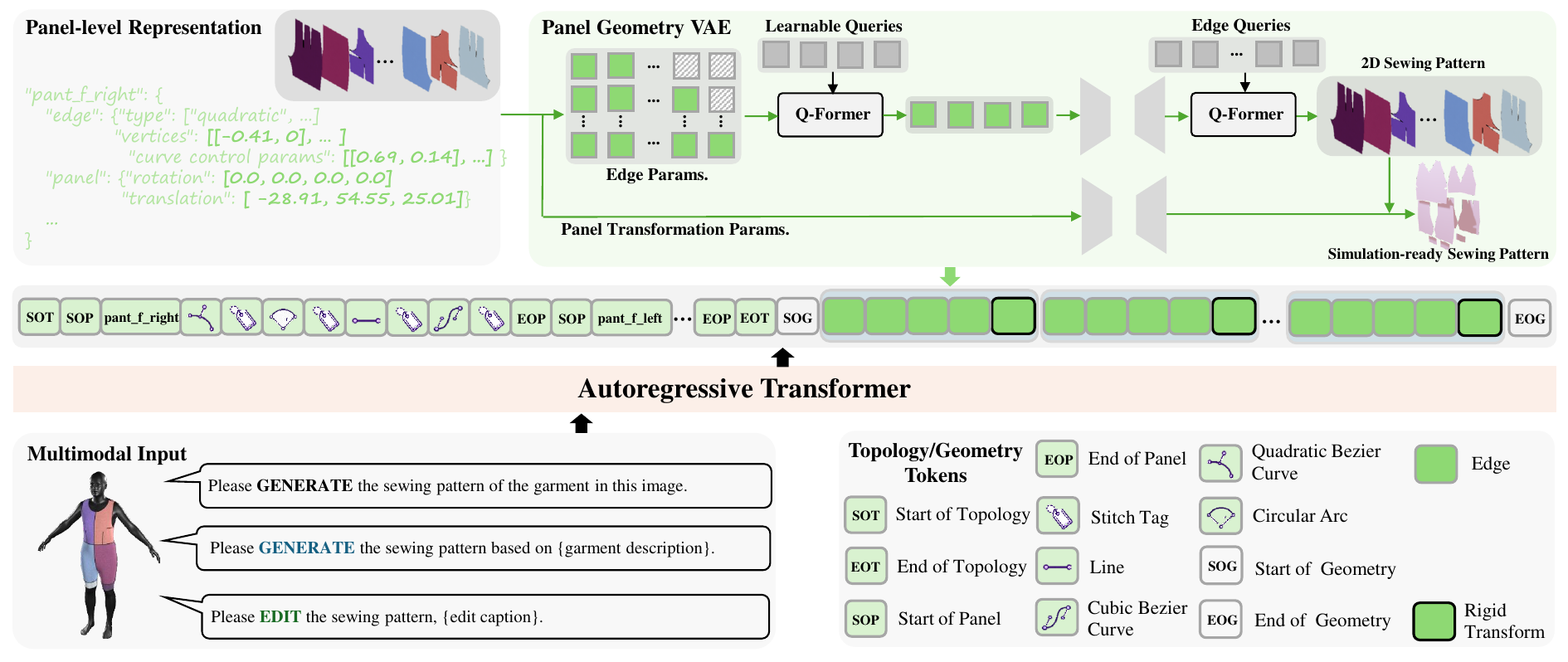}
    \caption{\textbf{Overview of SewFusion.} Our model comprises a Panel Geometry VAE and a Panel Geometry Flow. The Panel Geometry VAE compresses variable-length panel geometry and transformations into fixed-length latents and reconstructs continuous geometry. Conditioned on images, text, or editing instructions, a causal Transformer autoregressively generates sewing pattern topology and geometry tokens, whose hidden states guide flow-based latent sampling. The sampled geometry is then decoded together with the predicted topology to produce editable and simulation-ready sewing patterns.}
    \label{fig:overview}
    \vspace{-4mm}
\end{figure*}

\section{Method} 

\subsubsection{Overview.}
\label{sec:method_overview}
As illustrated in Fig.~\ref{fig:overview}, we propose SewFusion, a unified autoregressive framework that adopts tailored generation mechanisms for discrete topology and panel-level continuous geometry. SewFusion first generates discrete topology at the token level through next-token prediction, providing the structural condition for subsequent geometry generation. To support panel-level continuous geometry modeling, a Panel Geometry VAE encodes the variable-length geometry parameters of each panel into a fixed-size latent representation, which is then generated by Panel Geometry Flow conditioned on the multimodal input and generated topology. Panel-Forcing further exposes the geometry generator to model-generated topology during training, reducing the training--inference mismatch in topology context and improving robustness to topology prediction errors. We next introduce these components in order.

\subsection{Token-Level Discrete Topology Generation}

SewFusion employs a causal multimodal Transformer~\cite{liu2023visual} to generate discrete topology from the multimodal input $x$, which may include a garment image, a text instruction, and, for pattern editing, an existing sewing pattern. Following~\cite{nakayama2025aipparel}, we serialize the topology of the $i$-th panel as $T_i=(t_{i,1},\ldots,t_{i,L_i})$, where $L_i$ denotes the number of topology tokens in the panel and $t_{i,j}$ denotes the $j$-th token, such as a panel identity, edge type, or stitch tag. The complete topology $T=\{T_i\}_{i=1}^{N}$ is generated autoregressively and trained through next-token prediction:
\begin{equation}
\mathcal{L}_{\mathrm{topo}}
=
-\sum_{i=1}^{N}
\sum_{j=1}^{L_i}
\log p_{\theta}
\left(
t_{i,j}
\mid
T_{<i},t_{i,<j},x
\right),
\label{eq:topology_loss}
\end{equation}
where $T_{<i}$ denotes the topology of previously generated panels and $t_{i,<j}$ denotes the preceding topology tokens within the current panel. The generated topology further configures subsequent continuous geometry generation by determining the number of panel-level latent groups and edge queries.

\subsection{Panel-Level Continuous Geometry Generation}

\subsubsection{Panel Geometry VAE.}

Given the discrete topology generated above, directly modeling the continuous geometry of each panel remains difficult, since edge geometry varies in both the number of edges and the dimensionality of type-specific curve control parameters. Thus, we introduce a Panel Geometry VAE that maps each panel into a fixed-size latent representation. Since edge geometry is variable-length whereas the panel transformation is fixed-dimensional, the VAE employs separate edge and transformation branches, producing $K$ edge latent slots and one transformation latent slot for each panel.

For the edge branch, we standardize each edge representation and use a Q-Former~\cite{zhang2024qformer} to encode the resulting variable-length sequence into $K$ latent slots. Specifically, we consider the $i$-th panel and omit the panel index for clarity. Let $n$ denote its number of edges and $j\in\{1,\ldots,n\}$ index the edges in boundary order. The parameters of the $j$-th edge are arranged into a unified vector $e_j\in\mathbb{R}^{7}$ according to a predefined layout, together with a validity mask $m_j\in\{0,1\}^{7}$ indicating the active dimensions. The input embedding of each edge is constructed as:
\begin{equation}
u_j
=
W(e_j\odot m_j)
+
e_{\mathrm{pos}}(j),
\end{equation}
where $W\in\mathbb{R}^{d\times 7}$ is a learnable linear projection, $e_{\mathrm{pos}}(j)\in\mathbb{R}^{d}$ is a learned positional embedding, and $u_j\in\mathbb{R}^{d}$ is the resulting edge embedding. These edge embeddings are used as the keys and values of a Q-Former, whose $K$ learned queries aggregate them into a fixed-size representation. Linear projections then predict the mean and log-variance of the edge latent distribution. During decoding, edge queries attend to the sampled latent slots to reconstruct the continuous parameters of the corresponding edges.

For the transformation branch, the fixed-dimensional transformation $H_i\in\mathbb{R}^{7}$, consisting of a translation $t_i\in\mathbb{R}^{3}$ and a unit quaternion $q_i\in\mathbb{R}^{4}$, is projected to predict the mean and log-variance of a single latent slot. The sampled latent is decoded to reconstruct $\hat{t}_i$ and $\hat{q}_i$, with $\hat{q}_i$ normalized to unit length before computing the reconstruction loss.

We separately pretrain the edge and transformation branches of the Panel Geometry VAE. The edge branch is optimized with masked reconstruction of valid edge parameters, auxiliary geometric supervision on curve control parameters and the resulting panel contour, and KL regularization of the edge latent distribution. The transformation branch is independently optimized with translation and quaternion-rotation reconstruction, together with KL regularization of the transformation latent distribution. Detailed formulations are provided in the appendix.

\subsubsection{Panel Geometry Flow.}

With the structure provided by the preceding token-level discrete topology generation, Panel Geometry Flow then generates the continuous geometry of each panel in the compact, fixed-size latent space learned by the Panel Geometry VAE. Specifically, the number of panels parsed from the generated topology sequence determines how many panel-level latent groups are instantiated for flow matching, while the number of edge-type tokens predicted within each panel determines how many edge queries are initialized by the VAE decoder. These panel-level latents are generated within the same causal multimodal Transformer used for topology generation, conditioned on the multimodal input and the complete generated topology.

{Formally, during training,} let $N$ denote the number of panels in the topology sequence. Each panel corresponds to one latent group containing $K$ edge latent slots and $1$ transformation latent slot. For the $i$-th panel, the pretrained Panel Geometry VAE encodes its ground-truth geometry into the latent group $Z_{1,i}$. We sample a Gaussian noise group $Z_{0,i}\sim\mathcal{N}(0,I)$ and a flow time $t\sim\mathcal{U}(0,1)$, and define the linear probability path:
\begin{equation}
Z_{t,i}
=
(1-t)Z_{0,i}
+
tZ_{1,i},
\qquad
V_i^{*}
=
Z_{1,i}
-
Z_{0,i}.
\label{eq:flow_path}
\end{equation}

The interpolated latent group $Z_{t,i}$ is projected into the Transformer hidden space and augmented with flow-time and panel-category embeddings to form the geometry tokens $G_{t,i}$:
\begin{equation}
G_{t,i}
=
WZ_{t,i}
+
f_t(t)
+
e_{\mathrm{panel}}(c_i),
\label{eq:geometry_token}
\end{equation}
where $W$ is a learnable projection, $f_t(t)$ denotes the flow-time embedding, and $e_{\mathrm{panel}}(c_i)$ is a learnable embedding indexed by the panel category $c_i$ predicted in the topology sequence. The geometry tokens of all panel-level latent groups are appended after the topology sequence and jointly processed in a single forward pass of the same causal multimodal Transformer to predict their velocities $\{\hat{V}_i\}_{i=1}^{N}$. Panel Geometry Flow is optimized by:
\begin{equation}
\mathcal{L}_{\mathrm{flow}}
=
\sum_{i=1}^{N}
\left\|
\hat{V}_i
-
V_i^{*}
\right\|_2^2.
\label{eq:flow_loss}
\end{equation}

{During inference,} each predicted panel is assigned an independent Gaussian noise latent group, and all panel-level latent groups are jointly evolved through the predicted velocity field until the final VAE latents are obtained. The resulting latents are decoded by the pretrained Panel Geometry VAE decoder. For each panel, the predicted number of edges determines the number of edge queries used to reconstruct its edge geometry, while the transformation latent is decoded into the corresponding transformation.

\subsubsection{Panel-Forcing.}

Panel Geometry Flow is conventionally trained with ground-truth topology context, \textit{i.e.}, teacher-forcing, whereas inference conditions on model-generated topology. This discrepancy leaves the geometry generator unprepared for topology prediction errors at inference and may consequently degrade the generated panel geometry. Inspired by self-forcing~\cite{huang2025self}, we introduce Panel-Forcing to reduce this mismatch.

Specifically, the model first autoregressively generates a topology $\tilde{T}$ from the multimodal input $x$. The predicted topology $\tilde{T}$ is then used in place of the ground-truth topology $T$ as the conditioning context for Panel Geometry Flow, while the flow-matching targets remain derived from the ground-truth panel geometry. The resulting objective is
\begin{equation}
\mathcal{L}_{\mathrm{PF}}
=
\sum_{i=1}^{N}
\left\|
\hat{V}_{i}(\tilde{T},x)
-
V_i^{*}
\right\|_2^2,
\label{eq:panel_forcing}
\end{equation}
where $\hat{V}_{i}(\tilde{T},x)$ denotes the velocity predicted for the $i$-th panel under the model-generated topology context. By exposing Panel Geometry Flow to self-generated topology during training, Panel-Forcing improves its robustness to topology prediction errors encountered at inference.

\subsection{Training Strategy}

The edge and transformation branches of the Panel Geometry VAE are pretrained using their respective objectives and then kept fixed. Subsequently, the topology generator and Panel Geometry Flow are trained in two stages. During warm-up, both components use teacher-forced topology. After warm-up, the topology generator remains teacher-forced, while Panel Geometry Flow switches to Panel-Forcing and conditions on autoregressively generated topology.

\section{Experiment}

\begin{table*}[t]
    \centering
     \setlength{\abovecaptionskip}{0pt}
\caption{\textbf{Quantitative comparison with previous state-of-the-art methods on SewFactory and GCD-MM.}}
    \label{tab:quantitative_comparison}
    \setlength{\tabcolsep}{2.8pt}
    \renewcommand{\arraystretch}{1.08}

    \resizebox{\textwidth}{!}{%
    \begin{tabular}{lllcccccccccc}
        \toprule
        \textbf{Dataset}
        & \textbf{Methods}
        & \textbf{Setting}
        & \makecell{\textbf{Panel}\\\textbf{Acc.} $\uparrow$}
        & \makecell{\textbf{Edge}\\\textbf{Acc.} $\uparrow$}
        & \makecell{\textbf{Stitch}\\\textbf{Acc.} $\uparrow$}
        & \makecell{\textbf{Curve Type}\\\textbf{Acc.} $\uparrow$}
        & \makecell{\textbf{Vertex}\\$\mathbf{L_2}$ $\downarrow$}
        & \makecell{\textbf{Transl.}\\$\mathbf{L_2}$ $\downarrow$}
        & \makecell{\textbf{Rot.}\\$\mathbf{L_2}$ $\downarrow$}
        & \makecell{\textbf{Quad. Ctrl.}\\$\mathbf{L_2}$ $\downarrow$}
        & \makecell{\textbf{Cubic Ctrl.}\\$\mathbf{L_2}$ $\downarrow$}
        & \makecell{\textbf{Arc Ctrl.}\\$\mathbf{L_2}$ $\downarrow$} \\
        \midrule


        \multirow{3}{*}{SewFactory}
        & SewFormer
        & \multirow{3}{*}{\textit{image}}
        & 89.60
        & 99.06
        & 81.06
        & 90.53
        & 4.5921
        & 2.1657
        & 0.0857
        & 0.1935
        & --
        & -- \\

        & AIpparel
        &
        & 98.43
        & 98.43
        & 88.62
        & 99.88
        & 3.1488
        & 1.5769
        & 0.0275
        & 0.0951
        & --
        & -- \\

        & \textbf{SewFusion (Ours)}
        &
        & \textbf{98.66}
        & \textbf{99.14}
        & \textbf{90.88}
        & \textbf{99.88}
        & \textbf{2.6654}
        & \textbf{1.0465}
        & \textbf{0.0231}
        & \textbf{0.0842}
        & --
        & -- \\

        \midrule


        \multirow{12}{*}{GCD-MM}
        & SewFormer
        & \multirow{4}{*}{\textit{image}}
        & 68.96
        & 47.60
        & 36.29
        & 45.30
        & 12.5066
        & 5.2676
        & 0.0126
        & 0.5365
        & 0.5782
        & 0.3209 \\

        & SewingLDM
        &
        & 76.75
        & 72.26
        & 71.58
        & 68.48
        & 6.0844
        & 3.4855
        & 0.4855
        & 0.2497
        & 0.2429
        & 0.2012 \\

        & AIpparel
        &
        & 86.28
        & 80.83
        & 73.04
        & 79.54
        & 5.5278
        & 2.9291
        & 0.0084
        & 0.2146
        & 0.2451
        & 0.1802 \\

        & \textbf{SewFusion (Ours)}
        &
        & \textbf{92.33}
        & \textbf{85.64}
        & \textbf{81.74}
        & \textbf{82.23}
        & \textbf{4.1251}
        & \textbf{2.6933}
        & \textbf{0.0043}
        & \textbf{0.0896}
        & \textbf{0.1534}
        & \textbf{0.0397} \\

        \cmidrule(lr){2-13}


        & SewingLDM
        & \multirow{3}{*}{\textit{text}}
        & 59.72
        & 56.66
        & 58.26
        & 53.86
        & 10.8272
        & 8.0690
        & 0.0316
        & 0.4483
        & 0.4635
        & 0.3999 \\

        & AIpparel
        &
        & 61.03
        & 56.66
        & 54.58
        & 55.82
        & 7.8853
        & 3.3294
        & 0.0109
        & 0.3763
        & 0.3608
        & 0.2767 \\

        & \textbf{SewFusion (Ours)}
        &
        & \textbf{68.57}
        & \textbf{65.40}
        & \textbf{59.40}
        & \textbf{59.39}
        & \textbf{6.8230}
        & \textbf{3.1491}
        & \textbf{0.0062}
        & \textbf{0.1008}
        & \textbf{0.0931}
        & \textbf{0.0868} \\

        \cmidrule(lr){2-13}


        & SewingLDM
        & \multirow{3}{*}{\textit{image-text}}
        & 77.43
        & 75.22
        & 76.58
        & 58.08
        & 6.4484
        & 4.9755
        & 0.0114
        & 0.3370
        & 0.3552
        & 0.2959 \\

        & AIpparel
        &
        & 87.69
        & 81.77
        & 73.88
        & 80.46
        & 5.4643
        & 2.8954
        & 0.0083
        & 0.2121
        & 0.2423
        & 0.1782 \\

        & \textbf{SewFusion (Ours)}
        &
        & \textbf{94.05}
        & \textbf{89.74}
        & \textbf{85.18}
        & \textbf{84.85}
        & \textbf{3.5687}
        & \textbf{2.5326}
        & \textbf{0.0043}
        & \textbf{0.0897}
        & \textbf{0.0506}
        & \textbf{0.0399} \\

        \cmidrule(lr){2-13}


        & AIpparel
        & \multirow{2}{*}{\textit{editing}}
        & 82.91
        & 92.07
        & 93.64
        & 95.84
        & 7.7974
        & 3.0720
        & 0.0088
        & 0.2250
        & 0.2571
        & 0.1890 \\

        & \textbf{SewFusion (Ours)}
        &
        & \textbf{89.73}
        & \textbf{97.32}
        & \textbf{97.60}
        & \textbf{97.39}
        & \textbf{6.8309}
        & \textbf{2.1655}
        & \textbf{0.0023}
        & \textbf{0.1135}
        & \textbf{0.1301}
        & \textbf{0.0939} \\

        \bottomrule
    \end{tabular}%
    }

     \vspace{-4mm}
\end{table*}

\subsection{Experimental Setting}
\subsubsection{Datasets.}
We train our models on the SewFactory~\cite{liu2023sewformer} and GCD-MM~\cite{nakayama2025aipparel} datasets. GCD-MM is a multimodal extension of GarmentCodeData~\cite{korosteleva2023garmentcode} and contains approximately 115K garment samples; we follow its official training, validation, and test splits. After preprocessing and filtering invalid samples, we retain 8,269 SewFactory garments and randomly split them into training, validation, and test sets at a ratio of 9:0.5:0.5.

\subsubsection{Implementation Details.}

Following AIpparel~\cite{nakayama2025aipparel}, we employ LLaVA-1.5-7B~\cite{liu2023visual} as the backbone. In line with the GCD-MM protocol, we nominally sample image, text, image+text, and editing examples at a ratio of $3{:}2{:}4{:}1$ during training. We train on 8 H100 GPUs for 20,000 optimizer steps with a total batch size of 320. The learning rate increases to $10^{-4}$ over the first 100 steps and then follows cosine decay to zero. For the SewFactory dataset, which contains only the image modality, we train the model exclusively on the image-conditioned task. 


\subsubsection{Metrics.}  
 For discrete structure, we report panel-count, edge-count, stitch, and curve-type accuracies. For continuous geometry, edge geometry is evaluated using Vertex L2, Quadratic Control L2, Cubic Control L2, and Arc Control L2, while panel rigid transformations are evaluated using Translation L2 and sign-invariant Rotation L2.

\subsubsection{Baselines.} We compare with state-of-the-art \textbf{open-source} sewing pattern generation models, including SewFormer~\cite{liu2023sewformer}, diffusion-based SewingLDM~\cite{liu2025sewingldm} and autoregressive-based AIpparel~\cite{nakayama2025aipparel}.

\subsection{Image-based Generation}
\subsubsection{Task Setup.} Given a single rendered garment image, the model generates a garment sewing pattern. Since SewFactory represents all edges as single-control-point Bézier curves without explicit cubic or arc annotations, we omit the corresponding metrics. We evaluate only SewFormer and AIpparel, as SewingLDM requires sample-specific body parameters unavailable in SewFactory.

\subsubsection{Quantitative and Qualitative Results.}
\begin{figure}[t]
    \centering
    \setlength{\abovecaptionskip}{0pt}
    \includegraphics[width=\columnwidth]{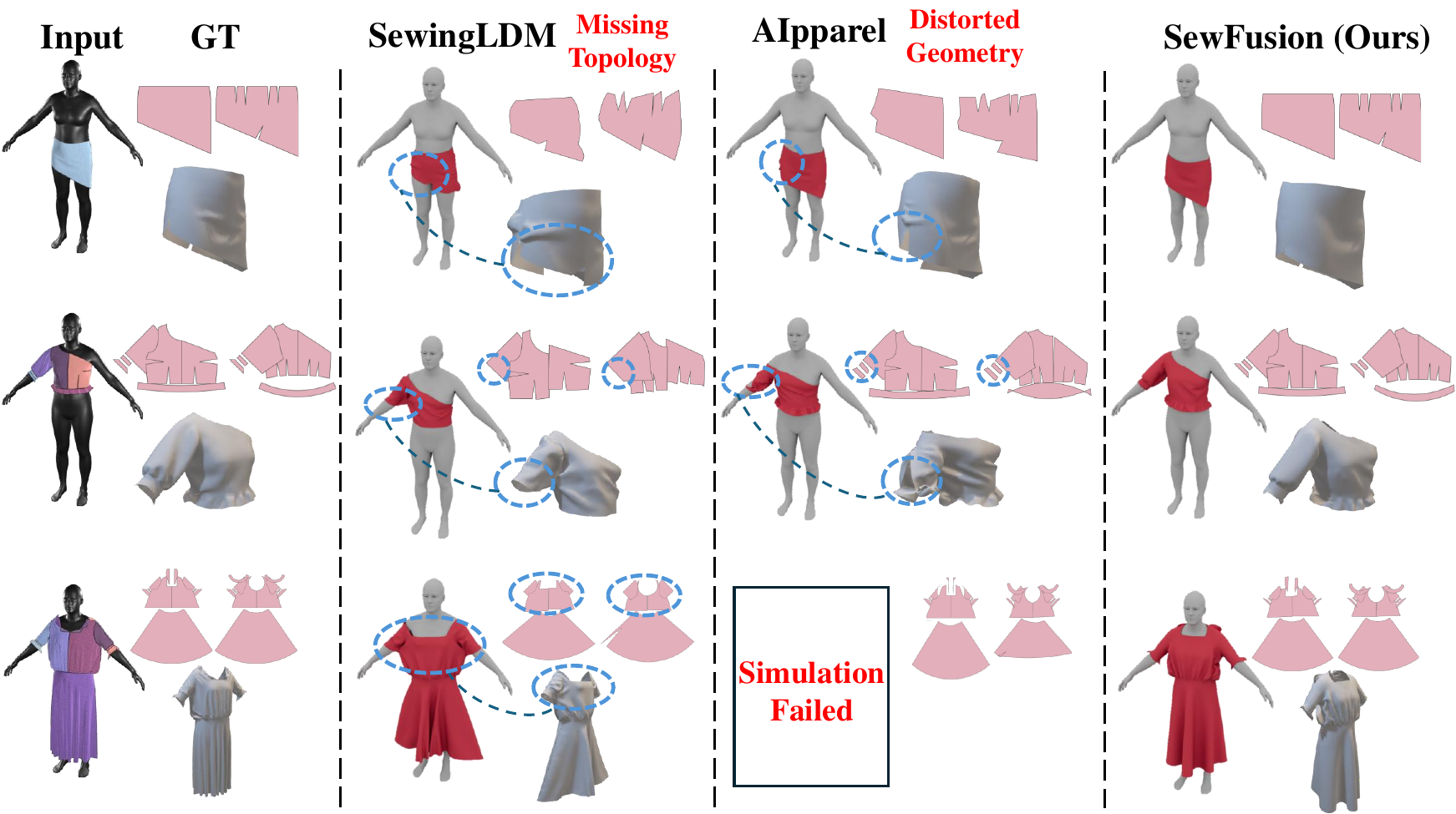}
    
    \caption{\textbf{Qualitative cases of image-based generation.}}
    \label{fig:image_qualitative}
\end{figure}

 Tab.~\ref{tab:quantitative_comparison} shows quantitative comparisons on the two datasets. Compared with the diffusion-based baseline, SewFusion achieves improvements of 15.58\% and 10.16\% in Panel Accuracy and Stitch Accuracy, respectively, together with a 0.1615 reduction in Arc control-point error. Compared with the token-level autoregression-based AIpparel, SewFusion improves Panel Accuracy and Stitch Accuracy by 6.05\% and 8.7\%, respectively, while reducing the curve control-point error by 0.1405. Fig.~\ref{fig:image_qualitative} shows qualitative results. SewingLDM predicted panel geometry violates the original topology, causing self-intersections that prevent simulation. AIpparel predicts sleeve geometry and stitching topology that deviate substantially from the ground truth, resulting in visible gaps around the arms. This suggests that baselines cannot adapt to complex garments with small panels and diverse edge types. In contrast, SewFusion predicts sewing pattern that match the input images, including small panels such as the waistband and the sleeve cuffs.

\subsection{Text-based Generation}
\subsubsection{Task Setup.} We evaluate three language-related conditioning modes on GCD-MM dataset. In \emph{text} generation, the input is a natural-language description of garment shape and style. 

\begin{figure}[t]
    \centering
     \setlength{\abovecaptionskip}{0pt}
    \includegraphics[width=\columnwidth]{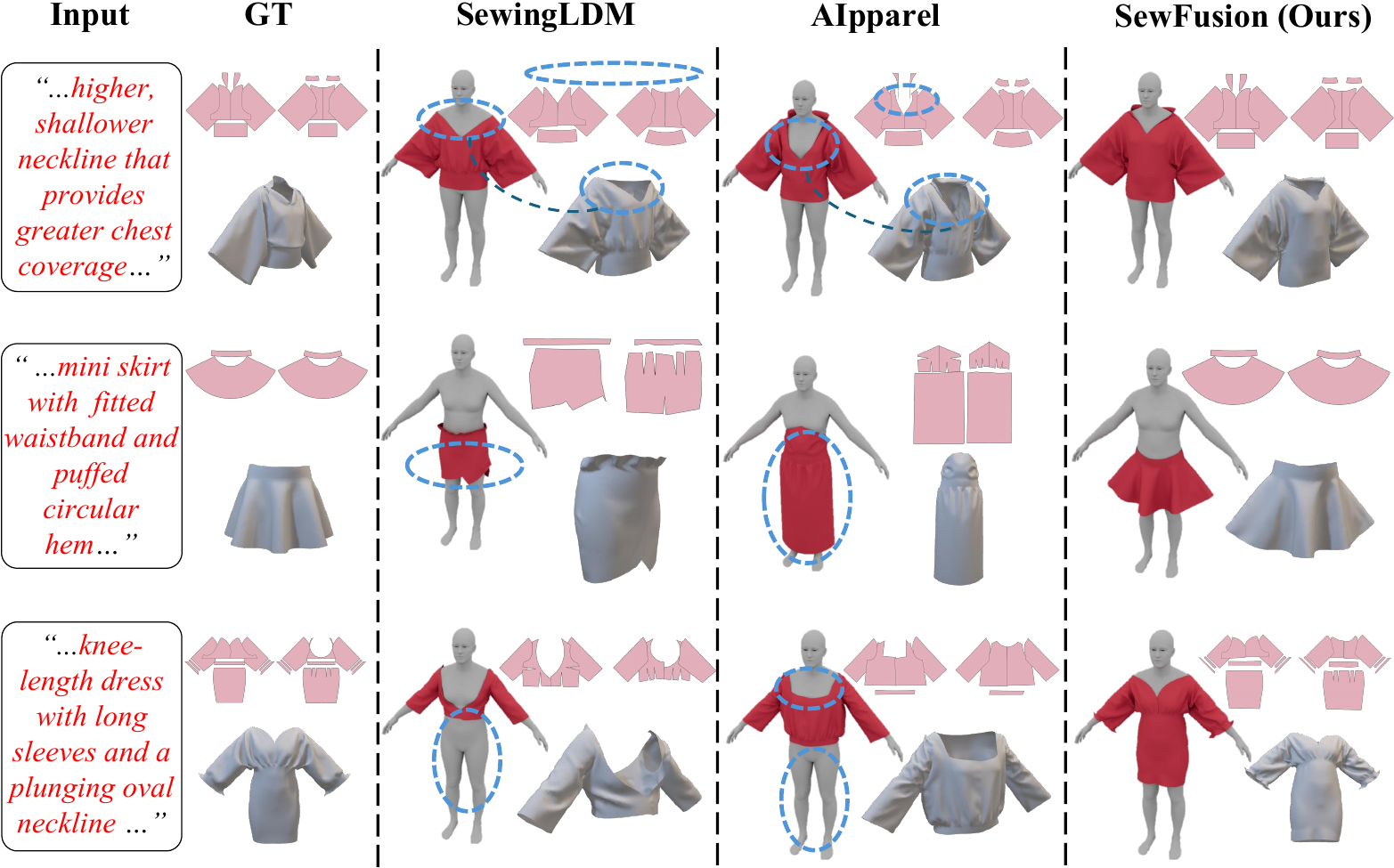}
   
    \caption{\textbf{Qualitative cases of text-based generation.} }
    \label{fig:text_example}
\end{figure}

\subsubsection{Quantitative and Qualitative Results.}
Tab.~\ref{tab:quantitative_comparison} shows quantitative comparisons on the GCD-MM dataset. Compared with the diffusion-based baseline, SewFusion achieves improvements of 8.85\% and 1.14\% in Panel Accuracy and Stitch Accuracy, respectively, together with a 0.3705 reduction in Cubic control-point error. Compared with the token-level autoregression-based AIpparel, SewFusion improves Panel Accuracy and Stitch Accuracy by 7.54\% and 4.82\%, respectively, while reducing the curve control-point error by 0.2677. Fig.~\ref{fig:text_example} shows qualitative results. SewingLDM and AIpparel often violate the requested global structure, producing incorrect necklines, skirt lengths, or garment categories. In contrast, SewFusion better follows the textual descriptions and generates sewing patterns with panel topology and geometry closer to the ground truth, demonstrating stronger text alignment and structural fidelity.

\subsection{Image-Text-based Generation}
\subsubsection{Task Setup.} In \emph{image-text} generation, a rendered garment image and its descriptive caption jointly condition the output.

\begin{figure}[t]
    \centering
    \setlength{\abovecaptionskip}{0pt}
    \includegraphics[width=\columnwidth]{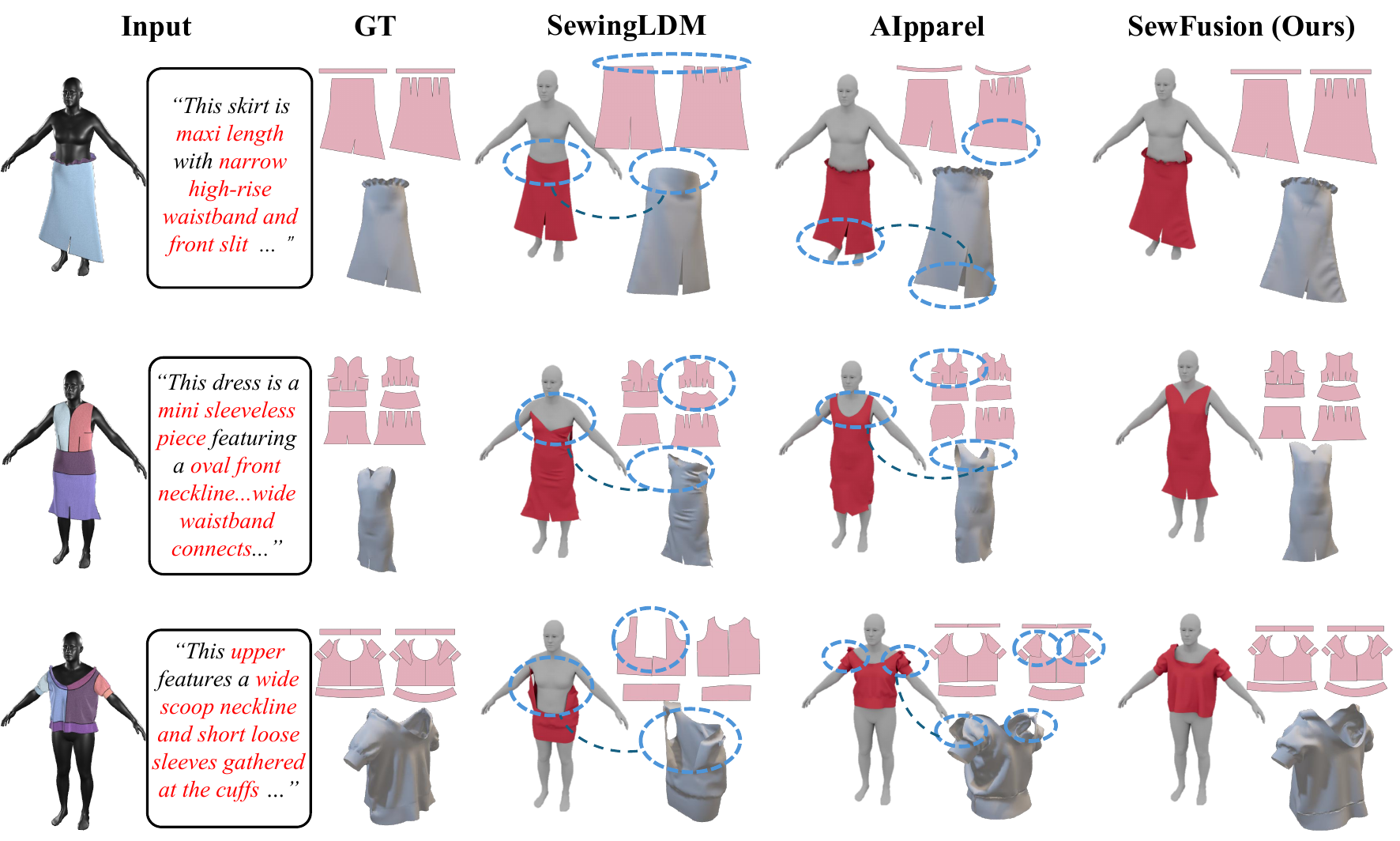}
    \caption{\textbf{Qualitative cases of image-text-based generation.}}
    \label{fig:text_image_example}
\end{figure}

\subsubsection{Quantitative and Qualitative Results.}
Tab.~\ref{tab:quantitative_comparison} shows quantitative comparisons on the GCD-MM dataset. Compared with the diffusion-based baseline, SewFusion achieves improvements of 16.62\% and 8.60\% in Panel Accuracy and Stitch Accuracy, respectively, together with a 0.3046 reduction in Cubic control-point error. Compared with the token-level autoregression-based AIpparel, SewFusion improves Panel Accuracy and Stitch Accuracy by 6.36\% and 11.3\%, respectively, while reducing the curve control-point error by 0.1917. Fig.~\ref{fig:text_image_example} shows qualitative results. SewingLDM and AIpparel often miss visual and textual details, such as waistbands, hemlines, necklines, and sleeves. In contrast, SewFusion better matches both modalities and produces panel topology and geometry that are more consistent with the ground truth.
\subsection{Sewing Pattern Editing}
\subsubsection{Task Setup.} Garment editing takes a source sewing pattern and a natural-language instruction as input and predicts a target pattern.

\begin{figure}[t]
    \centering
    \setlength{\abovecaptionskip}{0pt}
    \includegraphics[width=\columnwidth]{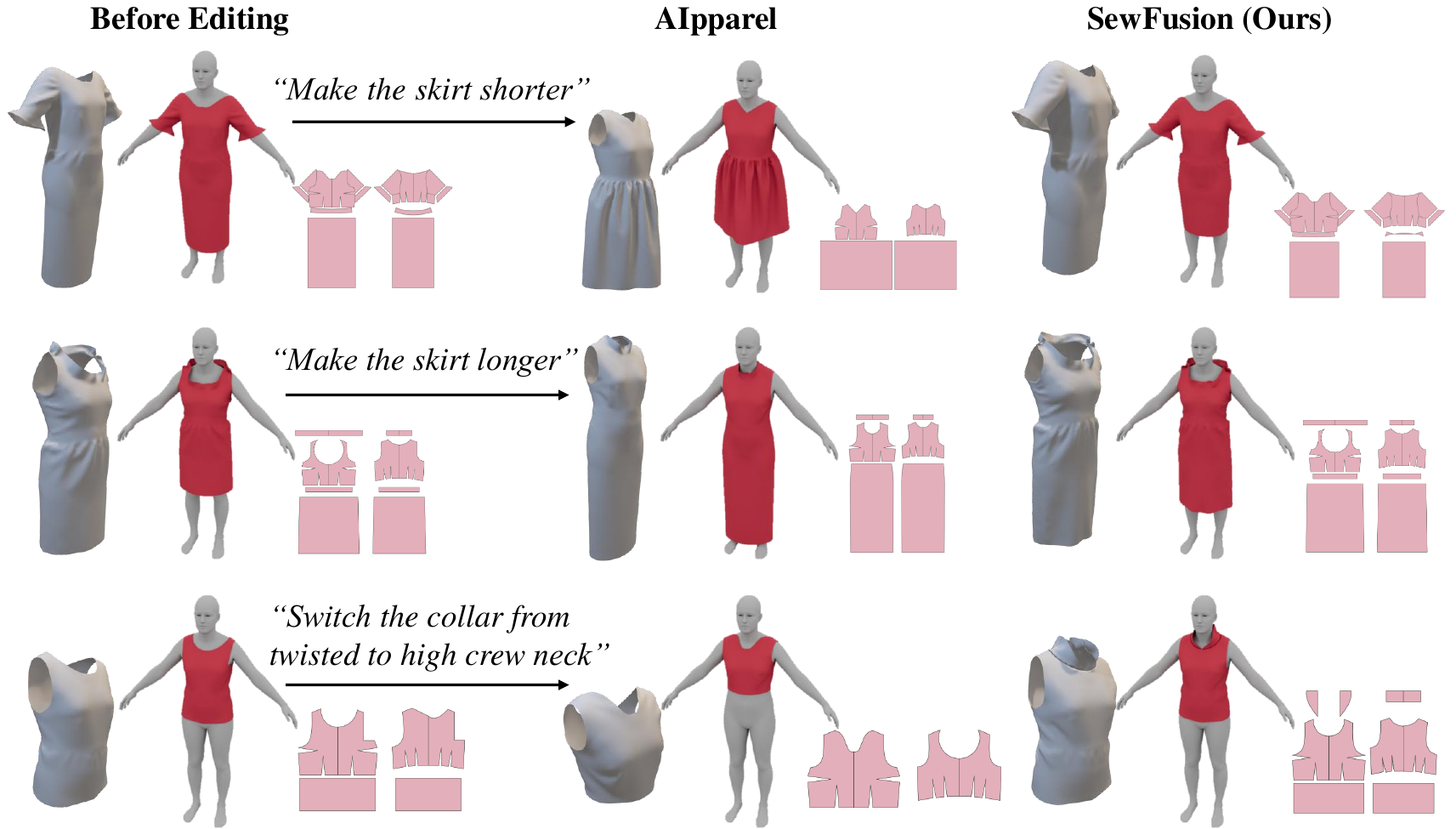}
    \caption{\textbf{Qualitative cases of sewing pattern editing.}}
    \label{fig:edit_example}
\end{figure}

\subsubsection{Quantitative and Qualitative Results.}
 Tab.~\ref{tab:quantitative_comparison} shows quantitative comparisons on the two datasets. SewFusion achieves superior performance on both metrics, demonstrating more precise edits while better preserving the unmodified regions of the sewing pattern. Fig.~\ref{fig:edit_example} shows qualitative results. AIpparel’s outputs exhibit noticeable deviations from the input garments. For example, it changes the short-sleeved maxi dress in the top row into a sleeveless dress with a different style, converts the square neckline in the middle row into a round neckline, and changes the round neckline in the bottom row into a V-neck. In contrast, by decoupling discrete topology from continuous geometry while explicitly modeling the topology, SewFusion provides more precise and controllable editing in response to the given instructions.

\begin{table}[t]

\setlength{\abovecaptionskip}{0pt}
\caption{\textbf{Ablation of Panel Geometry VAE token number.}}
\label{tab:vae_reconstruction}
\centering
\small
\setlength{\tabcolsep}{5.0pt}

\begin{tabular}{lcccc}
\toprule
Variant & Vertex L2 $\downarrow$ & Curve L2 $\downarrow$ & Transl. L2 $\downarrow$ & Rot. L2 $\downarrow$ \\
\midrule
$K=2$ & 0.0601 & 0.0322 & 0.0858 & 4.7e-06 \\
$K=4$ & 0.0358 & 0.0271 & 0.0725 & 4.0e-06 \\
$K=6$ & 0.0353 & 0.0272 & 0.0718 & 3.9e-06 \\
\bottomrule
\end{tabular}
\end{table}

\begin{table}[t]

\setlength{\abovecaptionskip}{0pt}
\caption{\textbf{Ablation of Panel-Forcing.}}
\label{tab:panel_forcing}
\centering
\small
\setlength{\tabcolsep}{4pt}
\begin{tabular}{llcccccccccc}
\toprule
 & Vertex $\downarrow$ & Transl. $\downarrow$&Rot. $\downarrow$&
Quad. $\downarrow$& Cubic $\downarrow$& Arc $\downarrow$ \\
\midrule
TF& 6.9240 &3.9493&0.0083& 0.1393 & 0.2199 & 0.1499\\

\textbf{PF}& \textbf{5.2075} &\textbf{3.6242}&\textbf{0.0047}& \textbf{0.0932}  & \textbf{0.1439}  & \textbf{0.0488} \\

\bottomrule
\end{tabular}
\end{table}

\subsection{Ablation Studies}
\label{sec:ablation_studies}

\subsubsection{Panel Geometry VAE.}
We evaluate the continuous representation by reconstructing test patterns using the VAE posterior mean. Table~\ref{tab:vae_reconstruction} varies the number of latent slots $K\in\{2,4,6\}$. Vertex L2 decreases from 0.0601  to 0.0353, while Curve L2 decreases from 0.0322 to 0.0272. Although K=6 achieves the lowest Vertex, Transl. and Rot. errors, K=4 gives a slightly lower Curve L2 and
requires fewer geometry tokens. To balance accuracy and efficiency, we set K=4 in the Panel Geometry VAE.

\begin{figure}[h!]
    \centering
    \setlength{\abovecaptionskip}{0pt}
    \includegraphics[width=\columnwidth]{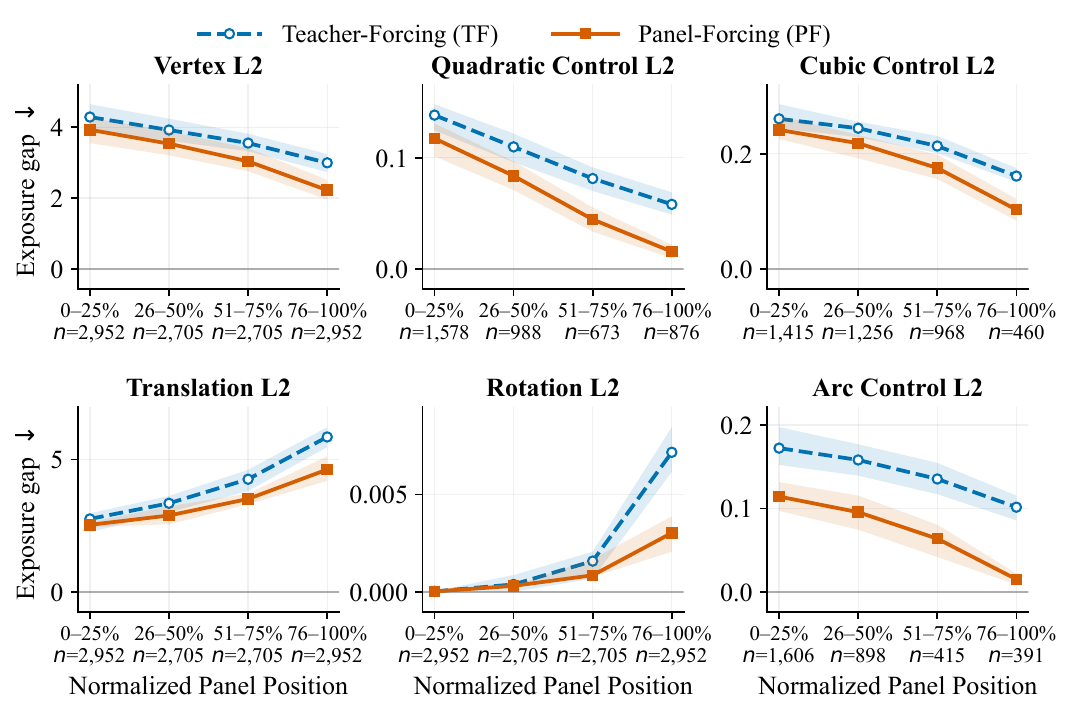}
    \caption{\textbf{Ablation of the Panel-Forcing exposure gap.}}
    \label{fig:fig_self_forcing_robustness}
\end{figure}

\subsubsection{Panel-Forcing.}
Because garments contain different numbers of panels, we normalize panel positions into four quartiles rather than using absolute indices. On GCD-MM, cubic curves are concentrated in early quartiles, so we report curve-specific exposure gaps to separate long-rollout degradation from curve-type imbalance. For a geometric metric $d$, the exposure gap is defined as $\Delta_{\mathrm{exp}}(d)=d(\tilde{S}) - d(S)$,
where $S$ and $\tilde{S}$ denote the ground-truth and generated topologies, respectively; positive values indicate degradation caused by autoregressive skeleton errors. Fig.~\ref{fig:fig_self_forcing_robustness} shows that Panel-Forcing consistently reduces these gaps, with larger improvements at later positions. From the first to the final quartile, the relative reduction increases from 8.4\% to 25.7\% for Vertex L2 and from 10.6\% to 57.8\% for Rotation L2, while final-quartile reductions reach 73.1\% and 85.1\% for Quadratic and Arc Control L2. Table~\ref{tab:panel_forcing} further shows that Panel-Forcing reduces Vertex L2 from 6.9240 to 5.2075 and the curve control-point error from 0.1499 to 0.0488, demonstrating improved robustness to accumulated autoregressive errors.


\section{Conclusion}
We introduced SewFusion, a hybrid autoregressive framework that preserves an explicit sewing pattern topology while generating continuous panel geometry in a structured latent space. Panel Geometry VAE represents boundary geometry and rigid transforms, which are generated through Panel Geometry Flow and decoded into editable vector patterns. Panel-Forcing further exposes the geometry predictor to model-generated topology contexts. Experiments on SewFactory and GCD-MM indicate improved continuous geometry without sacrificing explicit topology, while the ablation study shows that Panel-Forcing consistently reduces exposure gaps, particularly at later autoregressive positions.
\subsubsection{Limitations and Future Work.} Despite the improvements, several challenges remain to be addressed, including further improving simulation success rates, ensuring reliable sewability and physical plausibility, and generalizing robustly to more complex in-the-wild garment images.

\bibliography{aaai2027}


\end{document}